%% file: main.tex
\documentclass[letterpaper, 10 pt, conference]{ieeeconf}  %

\IEEEoverridecommandlockouts                              %

\usepackage{graphicx}
\usepackage{amsmath}
\usepackage{bbm}
\usepackage{dsfont}
\usepackage{subcaption}

\PassOptionsToPackage{dvipsnames}{xcolor}
\PassOptionsToPackage{table}{xcolor}

\usepackage{amssymb}
\usepackage{booktabs}
\usepackage{makecell}
\usepackage[ruled,noend,linesnumbered]{algorithm2e}
\usepackage{titlesec}
\usepackage{wrapfig}
\usepackage{array}
\usepackage{multirow}
\usepackage{colortbl}
\usepackage{tabularx}

\usepackage{grffile}
\usepackage{xspace}
\usepackage{needspace}

\usepackage{comment} 
\usepackage{url}
\usepackage{bm}
\usepackage{paralist}

\usepackage{pifont}

\usepackage[belowskip=0pt,aboveskip=5pt,font=small]{caption}
\usepackage[belowskip=0pt,aboveskip=0pt,font=small]{subcaption}
\usepackage{xcolor}
\PassOptionsToPackage{hyphens}{url}
\usepackage[hidelinks]{hyperref}
\usepackage{xfrac}
\usepackage{cite}

\usepackage{pifont} 
\definecolor{red}{rgb}{0.95,0.4,0.4}
\definecolor{blue}{rgb}{0.4,0.4,0.95}
\definecolor{darkblue}{rgb}{0,0,0.8}
\definecolor{darkred}{rgb}{0.8,0,0}
\definecolor{darkgreen}{rgb}{0.15,0.6,0.15}
\definecolor{grey}{rgb}{0.6,0.6,0.6}
\definecolor{col1}{RGB}{232, 161, 148}
\definecolor{col2}{RGB}{148, 187, 232}
\definecolor{tgre}{rgb}{0.15,0.6,0.15}
\definecolor{tred}{rgb}{0.6,0.15,0.15}
\newlength\savewidth

\title{Manydepth2: Motion-Aware Self-Supervised Monocular \\ Depth Estimation in Dynamic Scenes}

\author{
    Kaichen Zhou$^{1}$, 
    Jia-Wang Bian$^{1}$,
    Jian-Qing Zheng$^{1}$,
    Jiaxing Zhong$^{1}$, 
    Qian Xie$^{1}$,
    Niki Trigoni$^{1}$,
    Andrew Markham$^{1}$}

\begin{document}
\maketitle

\footnotetext[1]{KZ, JB, JZ, QX, NT and AM are with the University of Oxford. Contact Email: {\tt\small zhouk777@mit.edu}}

\thispagestyle{empty}
\pagestyle{empty}

\begin{abstract}
Despite advancements in self-supervised monocular depth estimation, challenges persist in dynamic scenarios due to the dependence on assumptions about a static world. 
In this paper, we present Manydepth2, to achieve precise depth estimation for both dynamic objects and static backgrounds, all while maintaining computational efficiency.
To tackle the challenges posed by dynamic content, we incorporate optical flow and coarse monocular depth to create a pseudo-static reference frame. 
This frame is then utilized to build a motion-aware cost volume in collaboration with the vanilla target frame. 
Furthermore, to improve the accuracy and robustness of the network architecture, we propose an attention-based depth network that effectively integrates information from feature maps at different resolutions by incorporating both channel and non-local attention mechanisms.
Compared to methods with similar computational costs, Manydepth2 achieves a significant reduction of approximately five percent in root-mean-square error for self-supervised monocular depth estimation on the KITTI-2015 dataset. 
The code could found \href{https://github.com/kaichen-z/Manydepth2}{\textcolor{blue}{https://github.com/kaichen-z/Manydepth2}}. 
\end{abstract}

\input{tex/1_intro}

\input{tex/2_related_work}

\input{tex/3_method}

\input{tex/4_experimental}

\newpage 
\newpage
\bibliographystyle{IEEEtran}
\bibliography{main}

\end{document}

%% file: tex/1_intro.tex
\section{Introduction}

The role of vision-based depth estimation (VDE) has become increasingly important in computer vision due to its ability to understand the 3D geometry of a scene based on 2D observation, which serves as the foundation for various high-level 3D tasks, such as scene reconstruction \cite{zhou2023dynpoint}, object detection \cite{li2021hierarchical} and navigation \cite{ye2021dpnet}. 
Moreover, VDE has enabled state-of-the-art applications ranging from autonomous driving \cite{dhamo2019peeking} to augmented reality \cite{dhamo2019peeking, zhou2023serf}.

Recently, self-supervised depth estimation has emerged as a viable approach for training depth estimation methods, aiming to alleviate the dependency on extensive training data and reduce high computational demands.
These methods learn depth maps from either monocular images \cite{bian2021unsupervised} or stereo image pairs \cite{yao2018mvsnet}. 
Despite the significant advancements made in self-supervised monocular vision-based depth estimation, a notable performance gap persists when comparing self-supervised monocular VDEs to self-supervised stereo VDEs.
 The disparity in performance can be mainly ascribed to the capability of stereo methods to utilize multiple views for constructing a feature volume, thereby incorporating a greater amount of 3D camera frustum information.
While multi-frame monocular VDEs, as presented in \cite{watson2021temporal}, have the ability to construct a feature volume based on adjacent frames, the presence of dynamic elements in these adjacent frames can potentially disrupt the construction of the feature volume.

\begin{figure}[htb]
    \centering
    \includegraphics[width=8.5cm]{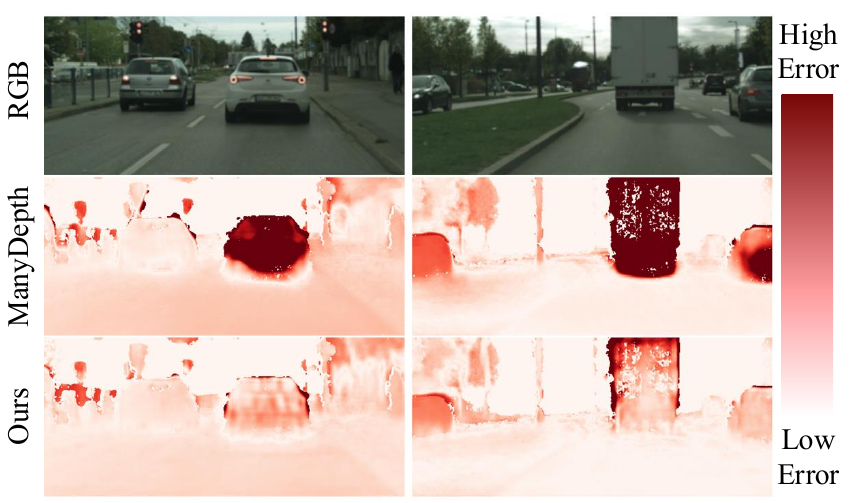}
    \caption{ \textbf{Qualitative comparison on Cityscapes.} The first row presents the RGB images. The second row shows the error maps of the depth estimates produced by ManyDepth. The third row displays the error maps of the depth estimates generated by Manydepth2.}
\label{fig:demo1}
\vspace{-0.4cm}
\end{figure} 

Taking the aforementioned concern into account, this paper strives to enhance the performance of multi-frame monocular VDEs by integrating motion information during the inference process and adeptly managing the influence of dynamic objects in the construction of the cost volume. 
Experimental results indicate that our method could effectively handle dynamic objects, as demonstrated in Fig.~\ref{fig:demo1}. 

Specifically, in light of recent advancements in optical flow estimation and its successful application in motion detection and estimation tasks, we propose Manydepth2. Manydepth2 is a self-supervised monocular vision-based depth estimation system that leverages motion information through a motion-aware cost volume constructed with an attention mechanism. The attention mechanism is chosen due to its demonstrated outstanding performance in representation learning and effective fusion of diverse information. Our main contributions are summarized as follows:

\begin{figure*}[htb]
    \centering
    \includegraphics[width=15.0cm]{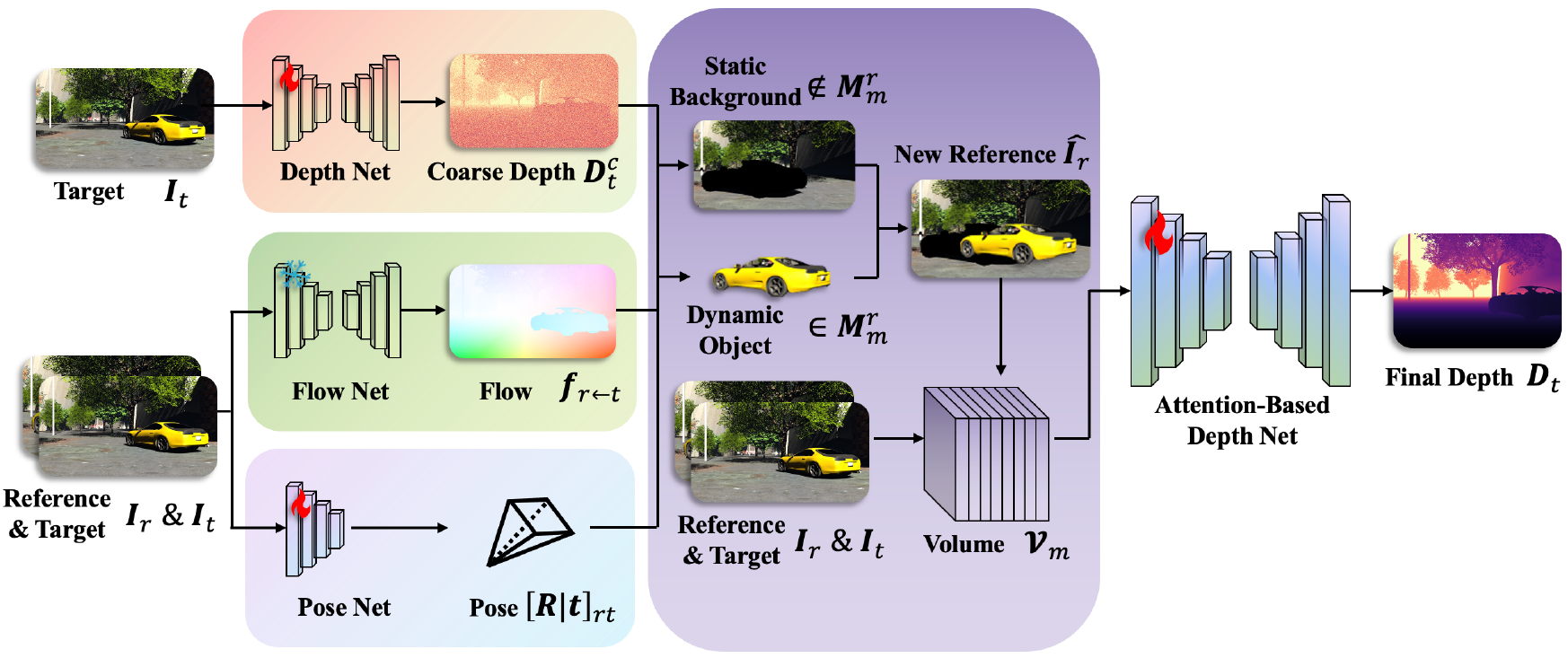}
    \caption{ \textbf{Illustration for the structure of Manydepth2.} During Stage 1, target and reference frame $\bm{I}_t$ and $\bm{I}_r$ are processed Flow Net, Pose Net, and Depth Net to generate optical flow $\bm{f}_{r\leftarrow t}$, transformation matrix $[\bm{R}|\bm{t}]_{rt}$, and coarse depth $\bm{D}^c_t$. During Stage 2, the outputs are used to generate the motion-aware cost volume $\bm{\mathcal{V}}_m$. Finally, the motion-aware cost volume $\bm{\mathcal{V}}_m$ and the target frame $\bm{I}_t$ are used by Attention-Based Depth Net to produce the refined depth $\bm{D}_t$.}
\label{fig:framework}
\vspace{-0.4cm}
\end{figure*} 

\begin{itemize}
    \item We utilize estimated optical flow alongside prior depth information to generate a pseudo-static reference frame. This reference frame effectively neutralizes the influence of dynamic elements within the original frame.

    \item By incorporating the pseudo-static reference frame, the target frame, and the initial reference frame, we construct a novel motion-aware volume that captures the dynamics of moving objects.
    
    \item Building on the High-Resolution Network (HRNet), we propose a novel depth estimation architecture that combines non-local and channel attention, enabling the integration of multi-scale features for accurate pixel-wise dense predictions.
    
    \item Our proposed model outperforms existing single and multi-frame methods on the KITTI, Cityscapes, and Odometry datasets. Additionally, our model can be trained efficiently using only a single NVIDIA RTX 3090 graphics card within a reasonable timeframe.
\end{itemize}

%% file: tex/2_related_work.tex
\section{Related Works}
\subsection{Monocular Depth Estimation}
Self-supervised techniques for monocular depth prediction have gained notable traction due to their independence from labeled data and their versatility across various scenarios.
Broadly, monocular VDEs fall into two categories: one relies on the present frame for depth estimation, while the other employs multiple frames to achieve depth prediction. \textbf{Single-frame Monocular Depth Estimation}: Within the realm of depth estimation methodologies, the initial category has witnessed the emergence of notable frameworks that have set benchmarks in terms of reliability and performance. A prime example from this category is Monodepth, as highlighted in the study by \cite{zhou2017unsupervised}. This model distinguishes itself with a bifurcated network architecture: one part is dedicated explicitly to pose estimation, while another part focuses on depth estimation tasks.
A pivotal aspect of the Monodepth architecture lies in the collaborative synergy between these two networks. They operate by leveraging the warping connection intrinsic to depth and image transformations. This collaboration results in creating a photometric loss function, which is a key part of why it works so well.
Further elevating the landscape of depth estimation methodologies, DevNet, in \cite{zhou2022devnet}, takes a comprehensive approach by embedding 3D geometric consistency. 
\textbf{Multi-frame Monocular Depth Estimation}:
The premise of the second category rests on the notion that integrating temporal information during inference—by employing multiple neighboring frames as inputs—can enhance the accuracy of the final depth estimation.
Initially, this is accomplished by employing test-time refinement techniques \cite{casser2019depth}, along with recurrent neural networks (RNNs) as evidenced in studies like \cite{patil2020don}.
The test-time refinement method adopts a monocular strategy to leverage temporal data during testing, whereas the recurrent neural network integrates with a monocular depth estimation network to analyze continuous frame sequences. 
Nonetheless, models utilizing recurrent neural networks often entail high computational costs and lack a distinct geometric inference approach.
Recently, Manydepth \cite{watson2021temporal} and MonoRec \cite{wimbauer2021monorec} have made notable advancements in performance and real-time efficiency by incorporating cost volumes from stereo-matching tasks for geometric-based reasoning \cite{gu2020cascade}. These models rely on a photometric loss function, where temporally neighboring frames are mapped onto the current image plane using predetermined depth bins.
Within the cost volume framework, the inferred depth with the minimal value corresponds closely to the actual depth.
Nevertheless, these approaches are grounded in static assumptions regarding scenarios and struggle with dynamic foreground elements. To tackle this limitation, we introduce Manydepth2, a technique adept at managing dynamic foreground by integrating temporal data into the cost volume and implementing a motion-aware photometric loss function.

\subsection{Self-supervised VDE for Dynamic Objects}

Owing to the unique attributes of dynamic objects, there has been a focused effort among researchers to handle these objects distinctively throughout both the training and inference stages within self-supervised vision-based depth estimation (VDE) techniques.
In \cite{klingner2020self}, a deliberate decision was made to exclude dynamic objects from the analysis. This exclusion was aimed at preventing any potential interference with the optimization process, thereby ensuring the precision and accuracy of the depth estimation results. On the other hand, in  \cite{feng2022disentangling}, a more nuanced approach was adopted. Dynamic objects were first identified and segmented from the scene. Subsequent to this segmentation, these objects were treated distinctively within the framework of the photometric loss function. Such a strategy facilitates the integration of dynamic objects into the depth learning mechanism, ensuring that they contribute meaningfully without merely being sidelined or excluded. Furthermore, recent research efforts, as seen in \cite{lee2021learning}, have ventured into predicting motion information at the object level. This predictive capability is then harnessed to establish more refined constraints, thereby enhancing the efficacy of self-supervised depth learning methodologies.
Despite these advancements, it's noteworthy that many of these approaches either grapple with intensive computational requirements or demonstrate only marginal enhancements in performance metrics. Recognizing these challenges, our novel approach, termed Manydepth2, seeks to overcome these constraints. We introduce a motion-aware cost volume framework that seamlessly integrates forecasted optical flow insights, all achieved through a sophisticated attention mechanism.

%% file: tex/3_method.tex
\section{Methodology}
\subsection{Overview}

Firstly, as a foundational step, we utilize the potential of both optical flow information and pre-training depth data. 
By combining these insights, we construct a pseudo-static reference frame. This innovative approach ensures that the depth estimation process remains robust and relevant, even when confronted with dynamic elements within the scene.
Our methodology further integrates the pseudo-static reference frame alongside the original target frame and vanilla reference frame to create a carefully designed motion-aware cost volume.
This volume could improve the depth estimation by considering the detailed movement and space relationships in the scene. 
Finally, We are introducing a depth net structure that utilizes attention mechanisms for processing cost volume. 
This structure is meticulously crafted to effectively combine feature maps that come from different resolutions, all originating from a high-resolution network.

\subsection{Framework}
Our objective is to estimate the depth map ${\bm{D}_t}$ and the rigid transformation ${[\bm{R}| \bm{t}]_{rt}}$ between the target frame ${\bm{I}_t}$ and a reference frame ${\bm{I}_r}$. To create the constraint for self-supervision, we can first reconstruct target frame ${\bm{I}'_t}$ in the following manner:
\begin{equation}
    \bm{I}'_t = \phi(\bm{I}_r, \bm{D}_t, \bm{K}_t, \bm{K}_r, [\bm{R}| \bm{t}]_{rt}).
\label{eqn:photometric1}
\end{equation}
Here, ${\bm{K}_t, \bm{K}_r}$ represents the intrinsic matrices of the target and reference frames, and $\phi(.)$ denotes the projection process based on the image warping. The ultimate photometric loss function, denoted as $\mathcal{L}_p$, is computed as the discrepancy between the transformed image ${\bm{I}'_t}$ and the reference image ${\bm{I}_t}$. This loss function serves as the optimization objective for the proposed neural network.

The structure of Manydepth2 is demonstrated in Fig.~\ref{fig:framework}. 
At the outset, Manydepth2 employs a trained optical flow network denoted by $\bm{\theta}_f$ and a pose network denoted by $\bm{\theta}_p$ to predict the optical flow $\bm{f}_{r\leftarrow t}$ and the transformation matrix $[\bm{R}| \bm{t}]_{rt}$ from the target frame to reference frame. Concurrently, a prior depth network denoted by $\bm{\theta}_{cd}$ is utilized to estimate the coarse depth map $\bm{D}_t^c$ for the target frame.
Subsequently, the target frame denoted as $\bm{I}_t$, the reference frame represented by $\bm{I}_r$, the coarse depth map denoted as $\bm{D}_t$, the transformation matrix $[\bm{R}| \bm{t}]_{rt}$, and the optical flow $\bm{f}_{r\leftarrow t}$ are utilized collaboratively to construct a motion cost volume denoted as $\bm{\mathcal{V}}_m$. The motion-aware cost volume $\bm{\mathcal{V}}_m$ and the target frame $\bm{I}_t$ are used in unison as inputs to the attention-based depth network $\bm{\theta}_{ad}$ with the objective of predicting the final depth map $\bm{D}_t$. Subsequently, this predicted depth map $\bm{D}_t$ is utilized in the construction of the ultimate photometric loss denoted as $\bm{\mathcal{L}}_{p}$.

\begin{figure}[t]
\setlength{\abovecaptionskip}{0.2cm}
\setlength{\belowcaptionskip}{-0.1cm}
    \centering
    \includegraphics[width=7.0cm]{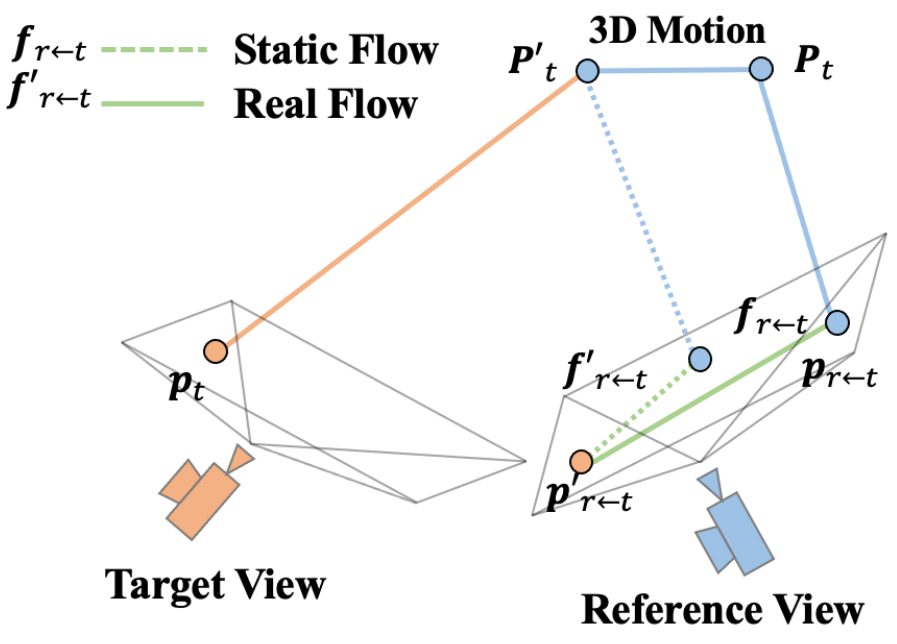}
    \caption{ \textbf{Relationship between optical flow and depth in the dynamic scenario.}  This figure demonstrates that in a dynamic scenario, there is a discrepancy between the static optical flow and the real optical flow.}
\label{fig:motion}
\vspace{-0.4cm}
\end{figure} 

\subsection{Pseudo-Static Reference Frame}
\label{sec:motion_estimate}

To address the depth estimation of moving objects with monocular video, we leverage the estimation of flow $\bm{f}_{r\leftarrow t}$, transformation matrix $[\bm{R}| \bm{t}]_{rt}$, and coarse depth $\bm{D}^c_t$.
Using the image warping relationship between $\bm{I}_t$ and $\bm{I}_r$, we can calculate a depth-based flow (static flow) $\bm{f}'_{r\leftarrow t}$ according to the following expression:
\begin{equation}
    \bm{f}'_{r\leftarrow t} = \frac{1}{\bm{D}_r}(\bm{K}_r)(\bm{R}_{rt}(\bm{K}_t)^{-1}\bm{D}^c_t p_t + \bm{t}_{rt}) - \bm{p}_t,
\label{eqn:movingflow}
\end{equation}
where $\bm{p}_t$ are pixels in the frame $\bm{I}_t$; $\bm{K}_r \& \bm{K}_t$ are intrinsic matrix for frames $\bm{I}_r \& \bm{I}_t$. 
As illustrated in Fig.~\ref{fig:motion}, in static scenarios, the static flow $\bm{f}'_{r\leftarrow t}$ aligns with the real optical flow $\bm{f}_{r\leftarrow t}$. 
In scenarios involving moving objects, the real optical flow $\bm{f}_{r\leftarrow t}$ can be decomposed into static optical flow $\bm{f}'_{r\leftarrow t}$ and dynamic optical flow $\bm{f}^d_{r\leftarrow t}$. 
Then the dynamic flow can be calculated as $\bm{f}^d_{r\leftarrow t} = \bm{f}_{r\leftarrow t} - \bm{f}'_{r\leftarrow t}$.

\begin{figure*}[t]
\vspace{-0.1cm}
\setlength{\abovecaptionskip}{0.1cm}
    \centering
    \includegraphics[width=17.0cm]{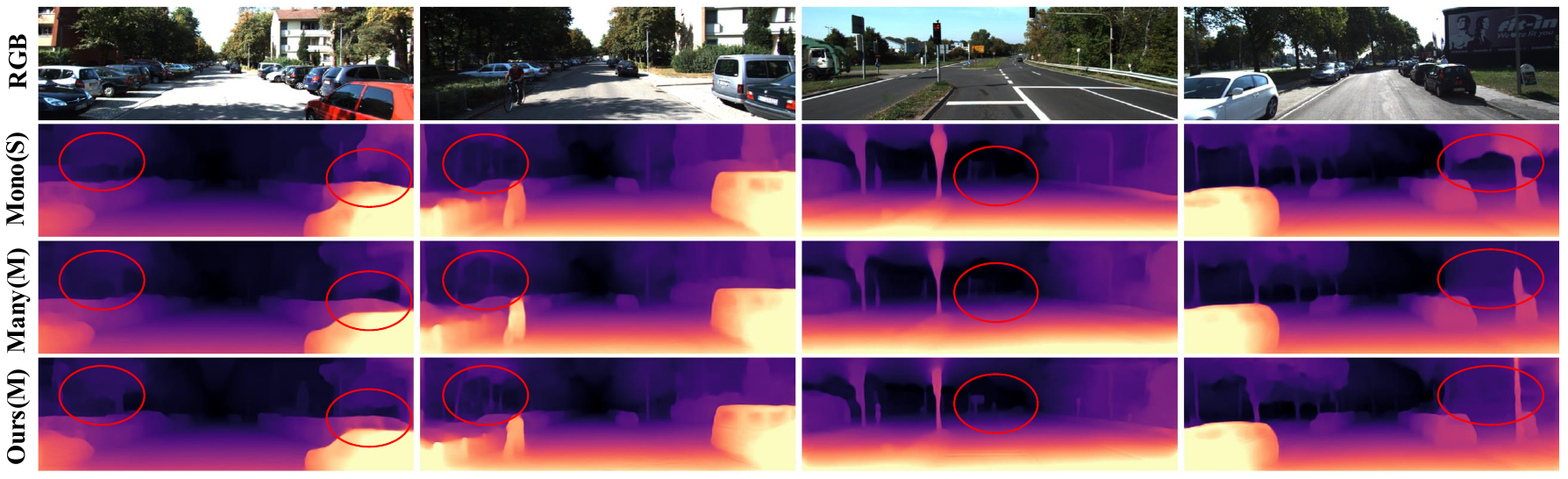}
    \caption{ \textbf{Qualitative results on KITTI Dataset.} The initial row displays the RGB images of target frames where depth information has been estimated. The subsequent rows, specifically the second, third, and fourth, showcase depth maps generated by Monodepth2 trained with stereo techniques, ManyDepth, and Manydepth2, respectively.}
\label{fig:demo6}
\vspace{-0.3cm}
\end{figure*} 

The motion mask could be generated as following:
\begin{equation}
\bm{\mathcal{M}}_m^t = ||\bm{f}_{r\leftarrow t} - \bm{f}'_{r\leftarrow t}||^2_2 > \epsilon,
\label{eqn:mask}
\end{equation}
where $\epsilon$ is the threshold for distinguishing moving parts. 
The motion mask in the target mask can be used to generate the motion mask $\bm{\mathcal{M}}_m^r$ in the reference mask, utilizing $\bm{f}_{r\leftarrow t}$. With $\bm{\mathcal{M}}_m^r$, it becomes possible to reconstruct a pseudo-static reference image $\hat{\bm{I}_r}$ by eliminating the impact of dynamic objects. This process can be formulated as follows: 
\begin{equation}
  \hat{\bm{I}_r} =
    \begin{cases}
      \bm{I}_r & \bm{p}_i \notin \bm{\mathcal{M}}_m^r ;\\
      \phi'(\bm{I}_t, \bm{D}_t, \bm{K}_t, \bm{K}_r, [\bm{R}| \bm{t}]_{rt}) & \bm{p}_i \in \bm{\mathcal{M}}_m^r .\\
    \end{cases}       
\end{equation}
It is important to note that $\phi'(.)$ differs from $\phi(.)$, as $\phi(.)$ is utilized to generate current frame information via sampling based on pixel correspondence between the current frame and the adjacent frame, whereas $\phi'(.)$ generates adjacent frame by shifting forward pixels of the current frame.

\subsection{motion-aware Cost Volume}

In constructing the motion-aware cost volume $\bm{\mathcal{V}}_m$ for target frame $\bm{I}_t$, a set of parallel planes that are perpendicular to the optic axis of $\bm{I}_t$ is defined, based on the depth assumptions $\bm{\mathcal{D}}=\{d_k; k=1,...,M\}$, where $M$ represents the number of planes. 
The feature extractor $\bm{\theta}_{fe}$ is utilized to generate the feature maps $\bm{\mathcal{F}}_t$ and $\hat{\bm{\mathcal{F}}_r}$ of $\bm{I}_t$ and $\alpha\hat{\bm{I}_r} + (1-\alpha)\bm{I}_r$, respectively. 
With the aid of $\bm{\mathcal{D}}$ and $[\bm{R}|\bm{t}]_{rt}$, a set of wrapped feature maps $\{\hat{\bm{\mathcal{F}}}_{t\leftarrow r}^{d_k}; d_k \in \bm{\mathcal{D}}\}$ are generated by warping $\hat{\bm{\mathcal{F}}_r}$. 
Considering $\bm{\mathcal{V}}_m = \{\bm{\mathcal{P}}_k, k=1,...,M\}$, it could be written as: 
\begin{equation}
    \bm{\mathcal{P}}_k = \frac{\sum_{i=1}^{N}| \hat{\bm{\mathcal{F}}}_{t\leftarrow r}^{d_k} - \bm{\mathcal{F}}_t|}{N}
\end{equation}

Here, $N$ denotes the number of reference frames used in construction, while in our experiment. $\alpha$ denotes the hyper-parameters used to balance the influence of pseudo-static reference frame $\hat{\bm{I}_r}$. 
Instead of directly taking the volume $\bm{\mathcal{V}}_m$ into the later depth network, we propose to construct the final feature volume $\bm{\mathcal{V}}_f$ by combining the $\bm{\mathcal{V}}_m$ with the $\bm{\mathcal{F}}_t$ though channel attention. The process of constructing the final feature volume $\bm{\mathcal{V}}_f$ is presented in Fig. \ref{fig:framework}.

\subsection{Attention-Based Depth Network}
The High-Resolution Network (HRNet) \cite{wang2020deep} is well-regarded for its ability to preserve a high level of detail in input images. 
The HRNet is composed of multiple branches, denoted as $B$, each generating $S = b, ..., 4$ features $\bm{f}^b_s$ with resolutions of $(\frac{H}{2^{b-1}}, \frac{W}{2^{b-1}})$. 
However, instead of exclusively utilizing the feature map from the last stage of each branch for disparity map prediction, we leverage the attention mechanism to integrate feature maps from the current branch's various stages and feature maps from deeper branches. 
To achieve fusion across both the channel and spatial dimensions, we introduce an innovative Depth Attention Network that leverages non-local attention $\mathcal{NA}$ and channel attention 
$\mathcal{CA}$.
Noting that the $j^{th}$ branch of depth decoder takes the $(B-j)^{th}$ branch as input.
This fusion process can be expressed as:
\begin{equation}
  \bm{x}^j =
    \begin{cases}
      \mathcal{CA}(\bm{f}^B_B, \quad [\bm{f}^{B-j-1}_s]_{s=B-j-1}^{B}) & j ==0 ;\\
      \mathcal{CA}(\mathcal{NA}(\bm{x}^j),  \quad [\bm{f}^{B-j-1}_s]_{s=B-j-1}^{B}) & j > 0 .\\
    \end{cases}       
\end{equation}

\subsection{Loss Function}
To summarize, there are three loss functions used in updating Manydepth2's weights. The final loss function $\mathcal{L}$ could be written as:
$ \mathcal{L} =  \mathcal{L}_p +  \mathcal{L}_s + \mathcal{L}_c. $
The photometric loss $\mathcal{L}_p$ consists of $L_1$ norm and SSIM regularization:
\begin{equation}
\begin{split}
    \mathcal{L}_p & = a \sum_{p} \bm{\mathcal{M}}_o \odot |\bm{I}_t - \bm{I}'_{t}| + b \frac{1 - \text{SSIM}(  \bm{I}_t,  \bm{I}'_{t})}{2},
\end{split}
\label{eqn:loss_photo}
\end{equation}
where $\bm{\mathcal{M}}_o$ is the auto mask introduced in \cite{godard2017unsupervised}. $\mathcal{L}_s$ is the smooth loss, which could be written as:
\begin{equation}
    \mathcal{L}_s = |\partial_x \bm{D}_t| e^{- |\partial_x \bm{I}_t|} + |\partial_y \bm{D}_t| e^{- |\partial_y \bm{I}_t|}.
\label{eqn:loss_smooth}
\end{equation}
And $\mathcal{L}_c$ is the consistency loss introduced by \cite{watson2021temporal}, to preserve the consistency between prior monocular depth and final multi-frame depth, which could be written as:
\begin{equation}
    \mathcal{L}_c = \sum \bm{\mathcal{M}} \odot |\bm{D}_t^c - \bm{D}_t|,
\end{equation}
where $\bm{\mathcal{M}}$ is the mask introduced in \cite{watson2021temporal}.

%% file: tex/4_experimental.tex
\section{Experiments}

\input{tex/table00}
\input{tex/table01}
The main tests and evaluations for our study were carried out using two primary datasets: KITTI-2015 \cite{geiger2012we} and Cityscapes \cite{silberman2012indoor}. In our experiments, the ResNet18 \cite{he2016deep} architecture was utilized for the pose network. The HRNet16 was used as the backbone for the attention-based depth network. The pre-trained Gmflow \cite{xu2022gmflow} on the Sintel dataset \cite{butler2012naturalistic} are used as optic flow network.
For testing, we set the minimum and maximum depth values to 0.1m and 80m, as in \cite{godard2019digging, jin2001real}.
In our experiment, the scale factor is computed using the median value of the ground-truth image as in \cite{zhou2017unsupervised}.

\subsection{Monocular Depth Estimation}
\noindent\textbf{KITTI-2015:} 
The Tab.~\ref{tab:kitti_eigen} presents a comparison between the performance of Manydepth2 and state-of-the-art algorithms.
Overall, Manydepth2 exhibits superior performance compared to other methods, irrespective of whether monocular or multiview images are used. 
In particular, Manydepth2 demonstrates a significant performance advantage over Manydepth, which also employs multiview images and a cost volume structure to estimate depth. Manydepth2 outperforms Manydepth by a substantial margin of $7.2\%$ in terms of $\text{Abs\ Rel}$.
In comparison to Dynamicdepth, which employs semantic motion segmentation maps to leverage high-level computer vision information for precise depth estimation, Manydepth2 achieves a superior performance of $5.3\%$ in terms of the Abs Rel, without the inclusion of any domain-specific information. Quantitative results are shown in Fig.~\ref{fig:demo6}. We also present the results for Manydepth2-NF, a version of Manydepth2 that does not use optical flow or the motion-guided cost volume. We found that our attention-based network outperformed Manydepth by $4.1\%$.

\noindent\textbf{Cityscapes:} 
In contrast to the KITTI dataset, the Cityscapes dataset features a higher percentage of dynamic objects. During the training phase on the Cityscapes dataset, the input resolution of Manydepth2 is set to $128\times416$. 
The lower section of Tab.~\ref{tab:city} indicates a substantial performance advantage of Manydepth2 over its competing methods.
Remarkably, Manydepth2 exhibits superior performance over ManyDepth, with a margin of approximately $15.0\%$ in the absolute relative error metric.
Through the visualization of error maps generated by computing the differences between predicted depth maps and their corresponding ground truth depth maps, Fig.~\ref{fig:demo1} facilitates a qualitative analysis of depth predictions. These observations indicate that Manydepth2 offers enhanced accuracy for depth estimation of both dynamic foreground and static background. Additionally, the incorporation of a pre-trained optical flow estimation model with low computational requirements has resulted in a minimal increase in parameter size and running time for Manydepth2. Quantitative results are shown in Fig.~\ref{fig:demo1}.
Tab.~\ref{tab:dynamic} showcases the depth errors within dynamic object regions by utilizing the ground truth motion segmentation maps provided in \cite{lee2021learning}, which corresponds to the qualitative results in Fig.~\ref{fig:demo1}.
\input{tex/table02}
\input{tex/table04}
\input{tex/table03}

\subsection{Odometry Estimation} %
\begin{figure}[htb]
\vspace{-0.1cm}
\setlength{\abovecaptionskip}{0.1cm}
    \centering
    \includegraphics[width=7.5cm]{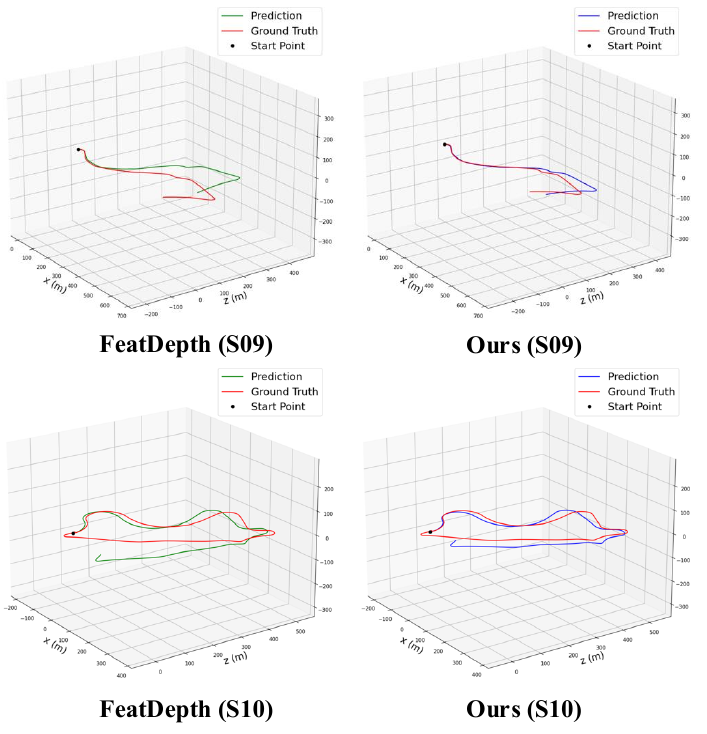}
    \caption{ \textbf{Trajectory visualization on Seq. 09. and Seq. 10.}  
    The ground truth trajectories are represented by red lines. Trajectories generated by the FeatDepth~\cite{shu2020feature} results are indicated by green lines, while our method's trajectories are depicted in blue lines.}
\label{fig:odom}
\vspace{-0.3cm}
\end{figure}

To evaluate odometry estimation results, we follow the split used in \cite{zhou2017unsupervised, zhan2018unsupervised}. Specifically, we trained our model using Seq. 00-08 from the KITTI odometry dataset and test methods on Seq. 09-10.
Tab.~\ref{tab:odom} shows the results,
where the translational and rotational motion were evaluated by using the root mean square error (RMSE).  Manydepth2, which uses less computational resources, outperforms other learning-based techniques that use more computational resources, such as FeatDepth.
The results indicate that Manydepth2 achieves a significant reduction in RMSE by approximately $24.1\%$ for translational motion and $22.5\%$ for rotational motion, as compared to ManyDepth, on Seq. 10. In support of this result, we present a visual depiction of the trajectory of both methods in Fig.~\ref{fig:odom}, which shows that Manydepth2's trajectory exhibits considerably less drift than that of FeatDepth.
The superior performance of Manydepth2 in precise pose estimation can be attributed to its use of both depth-based photometric loss and flow-based ego-motion loss, as well as the implementation of a motion-guided mask that effectively filters out dynamic foreground outliers. This combination of techniques enables Manydepth2 to achieve more reliable pose estimates compared to others.

\subsection{Abaltion Study}

In our research, we meticulously carry out ablation studies using the Cityscapes dataset. These studies are designed to dissect and showcase how effective our proposed modules are, allowing us to validate the specific design choices we've made. The detailed outcomes of these experiments are showcased in Table~\ref{tab:ablation}. Within this table, we provide insights from experiments conducted under diverse conditions and configurations. Specifically, we assess Manydepth2's performance with variations such as the Prior Depth Net, Manydepth2 configurations without Motion-Guided Volume and without Attention for Final Volume.
A key observation from Table~\ref{tab:ablation} is the pronounced positive impact that integrating the Attention-Based HRNet has on enhancing Manydepth2's performance. Furthermore, our findings underscore that the incorporation of both the Motion-Guided Volume and Attention for the Final Volume amplifies the capabilities of Manydepth2 even further, leading to improved outcomes.

\section{Conclusion}
We introduce Manydepth2, an innovative self-supervised multi-frame monocular depth prediction model that harnesses the synergy of optical flow-depth geometry. One of the standout features of Manydepth2 is its unique components, including a static reference frame, motion-guided cost volume, and an attention-enhanced high-resolution neural network. When benchmarked against the KITTI-2015 and Cityscapes datasets, Manydepth2 consistently delivers precise depth estimation, effectively distinguishing between dynamic foreground elements and static backgrounds. Remarkably, our model's performance is achieved using just a single RTX 3090, emphasizing its efficiency. Furthermore, Manydepth2 not only complements existing self-supervised monocular depth estimation techniques but also elevates them by integrating dynamic insights, thereby significantly enhancing accuracy.

\section{Acknowledgements} Our research is supported by Amazon Web Services in the Oxford-Singapore Human-Machine Collaboration Programme and by the ACE-OPS project (EP/S030832/1).

%% file: tex/table00.tex
{\newcommand{\x}{ x }
\newcommand{\midline}{  }
\newcommand{\splitline}{\arrayrulecolor{black}
\hhline{~-------------}}
\newcommand*\rot{\rotatebox{90}}
\definecolor{Asectioncolor}{RGB}{255,200,200}
\definecolor{Bsectioncolor}{RGB}{255,228,196}
\definecolor{Csectioncolor}{RGB}{235,255,235}
\definecolor{Dsectioncolor}{RGB}{235,235,255}
\begin{table*}[htb]
  \centering
  \resizebox{1.0\textwidth}{!}{
    \begin{tabular}{|l|c|l|c|c|c|c||c|c|c|c|c|c|c|}
        \arrayrulecolor{black}\hline
          &  T & Method & M  & S & F & W\x H & \cellcolor{col1}Abs Rel & \cellcolor{col1}Sq Rel & \cellcolor{col1}RMSE  & \cellcolor{col1}RMSE log & \cellcolor{col2}$\delta < 1.25 $ & \cellcolor{col2}$\delta < 1.25^{2}$ & \cellcolor{col2}$\delta < 1.25^{3}$ \\
         
        \hline\hline
        \parbox[b]{2mm}{\multirow{22}{*}{\rotatebox[origin=c]{90}{KITTI 2015}}} &  \parbox[b]{1mm}{\multirow{3}{*}{\rotatebox[origin=c]{90}{Stereo}}} & 
        EPC++\cite{luo2019every}&  & & & 1024\x320 & 0.128 & 0.935 & 5.011 & 0.209 & 0.831 & 0.945 & 0.979\\ 
        \midline
        & &FeatDepth\cite{shu2020feature} & &  & & 1024\x320 & 0.099 & 0.697 & 4.427 & 0.184 & 0.889 & 0.963 & 0.982 \\ 
        \midline
        & & DFR\cite{zhan2018unsupervised} & & & & 608\x160 & 0.135 & 1.132 & 5.585 & 0.229 & 0.820 & 0.933 & 0.971 \\ 
        \midline
        \cline{2-14}

        &  \parbox[b]{2mm}{\multirow{18}{*}{\rotatebox[origin=c]{90}{Monocular}}} & Packnet-SFM \cite{guizilini20203d}& & & &   1280\x384 &  0.107 & 0.802 & 4.538 & 0.186 & 0.889 & 0.962 & 0.981 \\
         \midline   
        & &Monodepth2  \cite{godard2019digging} & & &  & 1024\x320 &
         {0.115} &   {0.882} &   {4.701} &   {0.190} &   {0.879} &   {0.961} &   {0.982} \\ 
        \midline
        \midline
        \midline
         &  & FeatDepth \cite{shu2020feature} & & & & 1024\x320 & 0.104 & 0.729 & 4.481 & 0.179 & 0.893 & 0.965 & 0.984 \\
         \midline
        & & DevNet \cite{zhou2022devnet} &  & & &  1024\x320 & 0.103 & 0.713 & 4.459 & 0.177 & 0.890 & 0.965 & 0.982\\
        \midline
        & & Ranjan  \cite{ranjan2019competitive} &  & & $\bullet$ & 832\x256 & 0.148 & 1.149 & 5.464 & 0.226 & 0.815 & 0.935 & 0.973\\
        \midline
        & &  EPC++ \cite{luo2019every} & &  && 832\x256 & 0.141 & 1.029 & 5.350 & 0.216 & 0.816 & 0.941 & 0.976\\
        \midline
        &  &Guizilini  \cite{guizilini2020semantically}  & &$\bullet$ & & 640\x192 & 0.102 & 0.698 & 4.381 & 0.178 & 0.896 & 0.964 & 0.984 \\
        \midline
        & & Monodepth2 \cite{godard2019digging} &  & & &  640\x192 &
         {0.115} &   {0.903} &   {4.863} &   {0.193} &   {0.877} &   {0.959} &   {0.981} \\ 
        
        \midline
        & & Packnet-SFM  \cite{guizilini20203d}& & & & 640\x192 & 0.111 & 0.785 & 4.601 & 0.189 & 0.878 & 0.960 & 0.982 \\
        \midline
         & &Patil \cite{patil2020don} & $\bullet$ & $\bullet$ & &  640\x192 & 0.111 & 0.821 & 4.650 & 0.187 & 0.883 & 0.961 & 0.982\\        
        \midline
        \midline
         & &Li  \cite{li2023learning} & $\bullet$ & $\bullet$ & & 640\x192 & 0.102 & 0.703 & 4.348 & 0.175 & 0.895 & 0.966 & 0.984 \\
        \midline
        &  & ManyDepth\cite{watson2021temporal} &  $\bullet$  &  & & 640\x192  & 0.098 & 0.770 & 4.459 & 0.176 & 0.900 & 0.965 & 0.983 \\

        & & Lee\cite{lee2023multi} & $\bullet$ & & & 640\x192 & 0.096 & 0.644 & 4.230 & 0.172 & 0.903 & 0.968 & \textbf{0.985} \\

        & & AQUANet\cite{bello2024self} & $\bullet$ & & & 640\x192 & 0.105 & 0.621 & 4.227  & 0.179  & 0.889  & 0.964 & 0.984 \\
        
        & & DCPI-Depth\cite{zhang2024dcpi} & $\bullet$ & & $\bullet$ & 640\x192 & 0.095 & 0.662 & 4.274 & 0.170 & 0.902 & 0.967 & \textbf{0.985} \\
        
        & & DynamicDepth \cite{feng2022disentangling} & $\bullet$ & $\bullet$ & & 640\x192 & 0.096 & 0.720 & 4.458 & 0.175 & 0.897 & 0.964 &  0.984 \\     

        &  & \textbf{ManyDepth2-NF} &  $\bullet$  &  & & 640\x192  & 0.094 &  0.676 & 4.246 & 0.170 & \textbf{0.909} & \textbf{0.968} & \textbf{0.985} \\
        
        & & \textbf{{Manydepth2}} & $\bullet$ & & $\bullet$ & 640\x192 & \textbf{0.091} & \textbf{0.649} & \textbf{4.232} & \textbf{0.170} & \textbf{0.909} & \textbf{0.968} &  0.984 \\
        \arrayrulecolor{black}
        \hline
    \end{tabular}
  }  
  \caption{\textbf{Self-supervised depth estimation results on the KITTI 2015.}   
  We evaluate the performance of algorithms utilizing either stereo or monocular information during the training phase. 
  Metrics denoted by {\setlength{\fboxsep}{0pt}\colorbox{col1}{Color}} have a lower optimal value, while metrics denoted by {\setlength{\fboxsep}{0pt}\colorbox{col2}{Color}} have a higher optimal value.
  The optimal results in each subsection are denoted in \textbf{bold}. Our method outperforms all prior approaches in most metrics across all subsections, irrespective of whether the baselines utilize multiple frames during testing. M: Multi-frames. S: Motion Segmentation. F: Optical Flow. \textbf{Manydepth2-NF} refers to Manydepth2 without optical flow.}
\label{tab:kitti_eigen} 
\vspace{-0.2cm}
\end{table*}
}

%% file: tex/table01.tex
{\newcommand{\x}{ x }
\newcommand{\midline}{  }
\newcommand{\splitline}{\arrayrulecolor{black}
\hhline{~-------------}}
\newcommand*\rot{\rotatebox{90}}
\definecolor{Asectioncolor}{RGB}{255,200,200}
\definecolor{Bsectioncolor}{RGB}{255,228,196}
\definecolor{Csectioncolor}{RGB}{235,255,235}
\definecolor{Dsectioncolor}{RGB}{235,235,255}
\begin{table*}[htb]
  \centering
  \resizebox{1.0\textwidth}{!}{
    \begin{tabular}{|l|c|l|c|c|c|c||c|c|c|c|c|c|c|}
        \arrayrulecolor{black}\hline
          &  T & Method & M  & S & F & W\x H & \cellcolor{col1}Abs Rel & \cellcolor{col1}Sq Rel & \cellcolor{col1}RMSE  & \cellcolor{col1}RMSE log & \cellcolor{col2}$\delta < 1.25 $ & \cellcolor{col2}$\delta < 1.25^{2}$ & \cellcolor{col2}$\delta < 1.25^{3}$ \\
         
        \hline \hline
        \parbox[b]{2mm}{\multirow{11}{*}{\rotatebox[origin=c]{90}{Cityscapes}}} & \parbox[b]{2mm}{\multirow{11}{*}{\rotatebox[origin=c]{90}{Monocular}}} &
        Lee  \cite{lee2021attentive} &    & & & 832 \x 256  & 0.116 & 1.213 & 6.695 & 0.186 & 0.852 & 0.951 & 0.982 \\
        \midline
        & & InstaDM \cite{lee2021learning} &    & $\bullet$ & & 832 \x 256  & 0.111 & 1.158 & 6.437  & 0.182 & 0.868 & 0.961 & 0.983  \\
        \midline
        & &Pilzer \cite{pilzer2018unsupervised} &    & & & 512 \x 256  & 0.240 & 4.264 & 8.049 & 0.334 & 0.710 & 0.871 & 0.937  \\
        \midline
        & &Monodepth2 \cite{godard2019digging} &    & & & 416 \x 128  & 0.129 & 1.569 & 6.876 & 0.187 & 0.849 & 0.957 & 0.983  \\
        \midline
        & &Videos in the Wild \cite{gordon2019depth}&   & & & 416 \x 128   & 0.127 & 1.330 & 6.960 & 0.195 & 0.830 & 0.947 & 0.981  \\
        \midline
        & &Li  \cite{li2021unsupervised} &  & & & 416 \x 128   & 0.119 & 1.290 & 6.980 & 0.190 & 0.846 & 0.952 & 0.982  \\
        \midline
        & &Struct2Depth \cite{casser2019depth}& $\bullet$  & & & 416 \x 128 & 0.151 & 2.492 & 7.024 & 0.202 & 0.826 & 0.937 & 0.972  \\
        \midline
        &&ManyDepth\cite{watson2021temporal} & $\bullet$    & & & 416 \x 128  & 0.114 & 1.193 & 6.223 & 0.170 & 0.875 & 0.967 & 0.989 \\
        \midline
        & & DynamicDepth \cite{feng2022disentangling} & $\bullet$ & $\bullet$ & & 416 \x 128 & 0.103 & 1.000 & 5.867 & 0.157 & 0.895 & 0.974 &  0.991 \\     
        & &\textbf{Manydepth2} & $\bullet$    & & $\bullet$ & 416 \x 128   & \textbf{0.097} & \textbf{0.792} & \textbf{5.827} & \textbf{0.154} & \textbf{0.903} & \textbf{0.975} & \textbf{0.993} \\

        \arrayrulecolor{black}
        \hline
    \end{tabular}
  }  
  \caption{\textbf{Self-supervised depth estimation results on the Cityscapes.}   
  We evaluate the performance of algorithms utilizing either stereo or monocular information during the training phase. 
  Metrics denoted by {\setlength{\fboxsep}{0pt}\colorbox{col1}{Color}} have a lower optimal value, while metrics denoted by {\setlength{\fboxsep}{0pt}\colorbox{col2}{Color}} have a higher optimal value.
  The optimal results in each subsection are denoted in \textbf{bold}. Our method outperforms all prior approaches in most metrics across all subsections, irrespective of whether the baselines utilize multiple frames during testing. M: Multi-frames. S: Motion Segmentation. F: Optical Flow.}
\label{tab:city} 
\vspace{-0.2cm}
\end{table*}
}

%% file: tex/table02.tex
\newcommand{\splitlineblack}{\arrayrulecolor{black}\hhline{~-----------}}

\begin{table}[htb]
  \centering
  \resizebox{1.0\columnwidth}{!}{
  \begin{tabular}{|c|l|c|c||c|c|c|c|}
\arrayrulecolor{black}\hline
 Type & Method & M &  WxH  & \cellcolor{col1}Abs Rel & \cellcolor{col1}Sq Rel & \cellcolor{col1}RMSE  & \cellcolor{col1}RMSE log \\
  \hline
\arrayrulecolor{black}\hline\arrayrulecolor{gray}

S & InstaDM \cite{lee2021learning} &  & 832 x 256 & 
0.139 &  1.698 &  5.760 &  0.181   \\
\arrayrulecolor{black}\hline\hline
\multirow{3}{*}{w/o S} & Monodepth2 \cite{godard2019digging} &  &  416 x 128 & 0.159 &  1.937 &  6.363 &  0.201  \\ 
& ManyDepth \cite{watson2021temporal} & $\bullet$ &  416 x 128 & 0.169 &  2.175 &  6.634 &  0.218 \\  

& \textbf{Manydepth2} & $\bullet$ &  416 x 128 & \textbf{0.123} &  \textbf{1.260} &  \textbf{4.519} &  \textbf{0.144} \\ 
    \arrayrulecolor{black}\hline
  \end{tabular}
  }
  \caption{\textbf{Results of Dynamic Foreground on Cityscapes.} 
  We evaluate depth prediction errors for dynamic objects such as vehicles, people, and bikes on the Cityscapes dataset \cite{silberman2012indoor}. The best results are highlighted in bold. 
  }
  \label{tab:dynamic}
\vspace{-0.3cm}
\end{table}

%% file: tex/table04.tex
{\newcommand{\x}{ x }
\newcommand{\midline}{  }

\newcommand{\splitline}{\arrayrulecolor{black}\hhline{~------------}}

\newcommand*\rot{\rotatebox{90}}

\definecolor{Asectioncolor}{RGB}{255,200,200}
\definecolor{Bsectioncolor}{RGB}{255,228,196}
\definecolor{Csectioncolor}{RGB}{235,255,235}
\definecolor{Dsectioncolor}{RGB}{235,235,255}

\begin{table}[htb]
  \centering
  \footnotesize
  \resizebox{0.50\textwidth}{!}{
    \begin{tabular}{|l|c|c|c|c|}
        \arrayrulecolor{black}\hline
        Method & \cellcolor{orange!25}Tr & \cellcolor{orange!25}R & \cellcolor{green!25}Tr & \cellcolor{green!25}R   \\
        \hline\hline
        DFR \cite{zhan2018unsupervised}  & 11.93 & 3.91 & 12.45 & 3.46 \\ 
        NeuralBundler \cite{li2019pose} & 8.10  & 2.81 & 12.90 & 3.71 \\ 
        ManyDepth\cite{watson2021temporal} & 8.08 & 1.97 & 9.86 & 3.42  \\ 
        \textbf{Manydepth2} & \textbf{7.01} & \textbf{1.76} & \textbf{7.29} & \textbf{2.65} \\ 
        \arrayrulecolor{black}\hline
    \end{tabular}
  }  
  \vspace{-1pt}
  \caption{\textbf{Visual odometry results on {\setlength{\fboxsep}{0pt}\colorbox{orange!25}{Seq. 9}} and {\setlength{\fboxsep}{0pt}\colorbox{green!25}{Seq. 10}} of the KITTI odometry dataset.} 
  The average drift in root mean square error for both translation (Tr) and rotation (R) is reported.}
 \label{tab:odom} 
\vspace{-0.3cm}
\end{table}
}

%% file: tex/table03.tex
\begin{table}[htb]
  \centering
  \vspace{-2pt}
  \resizebox{1.0\columnwidth}{!}{
  \begin{tabular}{|l||c|c|c|}
  \hline
     Ablation & \cellcolor{col1}Abs Rel & \cellcolor{col1}Sq Rel & \cellcolor{col1}RMSE \\
  \hline\hline 
    Final Prior Depth & 0.105  &   0.768  &   4.536   \\
    w/o Attention for Volume & 0.100 & 0.709 & 4.298 \\
    Manydepth2-NF & 0.094 & 0.676 & 4.246  \\
    \textbf{Manydepth2} & \textbf{0.091} & \textbf{0.649} & \textbf{4.232}  \\
\hline
  \end{tabular}}
  \vspace{1pt}
  \caption{\textbf{Ablation study results of depth accuracy on the KITTI 2015.} The proposed Manydepth2 model was evaluated by modifying its components.}
  \label{tab:ablation}
  \vspace{-10pt}
\end{table}

%% file: main.bbl
\begin{thebibliography}{10}
\providecommand{\url}[1]{#1}
\csname url@rmstyle\endcsname
\providecommand{\newblock}{\relax}
\providecommand{\bibinfo}[2]{#2}
\providecommand\BIBentrySTDinterwordspacing{\spaceskip=0pt\relax}
\providecommand\BIBentryALTinterwordstretchfactor{4}
\providecommand\BIBentryALTinterwordspacing{\spaceskip=\fontdimen2\font plus
\BIBentryALTinterwordstretchfactor\fontdimen3\font minus \fontdimen4\font\relax}
\providecommand\BIBforeignlanguage[2]{{%
\expandafter\ifx\csname l@#1\endcsname\relax
\typeout{** WARNING: IEEEtran.bst: No hyphenation pattern has been}%
\typeout{** loaded for the language `#1'. Using the pattern for}%
\typeout{** the default language instead.}%
\else
\language=\csname l@#1\endcsname
\fi
#2}}

\bibitem{zhou2023dynpoint}
K.~Zhou, J.-X. Zhong, S.~Shin, K.~Lu, Y.~Yang, A.~Markham, and N.~Trigoni, ``Dynpoint: Dynamic neural point for view synthesis,'' \emph{Conference on neural information processing systems}, 2023.

\bibitem{li2021hierarchical}
S.~Li, J.~Shi, W.~Song, A.~Hao, and H.~Qin, ``Hierarchical object relationship constrained monocular depth estimation.'' \emph{Pattern Recognition}, vol. 120, p. 108116, 2021.

\bibitem{ye2021dpnet}
X.~Ye, S.~Chen, and R.~Xu, ``Dpnet: Detail-preserving network for high quality monocular depth estimation,'' \emph{Pattern Recognition}, vol. 109, p. 107578, 2021.

\bibitem{dhamo2019peeking}
H.~Dhamo, K.~Tateno, I.~Laina, N.~Navab, and F.~Tombari, ``Peeking behind objects: Layered depth prediction from a single image,'' \emph{Pattern Recognition Letters}, vol. 125, pp. 333--340, 2019.

\bibitem{zhou2023serf}
K.~Zhou, L.~Hong, E.~Xie, Y.~Yang, Z.~Li, and W.~Zhang, ``Serf: Fine-grained interactive 3d segmentation and editing with radiance fields,'' \emph{arXiv preprint arXiv:2312.15856}, 2023.

\bibitem{bian2021unsupervised}
J.-W. Bian, H.~Zhan, N.~Wang, Z.~Li, L.~Zhang, C.~Shen, M.-M. Cheng, and I.~Reid, ``Unsupervised scale-consistent depth learning from video,'' \emph{International Journal of Computer Vision}, vol. 129, no.~9, pp. 2548--2564, 2021.

\bibitem{yao2018mvsnet}
Y.~Yao, Z.~Luo, S.~Li, T.~Fang, and L.~Quan, ``Mvsnet: Depth inference for unstructured multi-view stereo,'' in \emph{Proceedings of the European conference on computer vision (ECCV)}, 2018, pp. 767--783.

\bibitem{watson2021temporal}
J.~Watson, O.~Mac~Aodha, V.~Prisacariu, G.~Brostow, and M.~Firman, ``The temporal opportunist: Self-supervised multi-frame monocular depth,'' in \emph{Proceedings of the IEEE/CVF Conference on Computer Vision and Pattern Recognition}, 2021, pp. 1164--1174.

\bibitem{zhou2017unsupervised}
T.~Zhou, M.~Brown, N.~Snavely, and D.~G. Lowe, ``Unsupervised learning of depth and ego-motion from video,'' in \emph{Proceedings of the IEEE conference on computer vision and pattern recognition}, 2017, pp. 1851--1858.

\bibitem{zhou2022devnet}
K.~Zhou, L.~Hong, C.~Chen, H.~Xu, C.~Ye, Q.~Hu, and Z.~Li, ``Devnet: Self-supervised monocular depth learning via density volume construction,'' in \emph{Computer Vision--ECCV 2022: 17th European Conference, Tel Aviv, Israel, October 23--27, 2022, Proceedings, Part XXXIX}.\hskip 1em plus 0.5em minus 0.4em\relax Springer, 2022, pp. 125--142.

\bibitem{casser2019depth}
V.~Casser, S.~Pirk, R.~Mahjourian, and A.~Angelova, ``Depth prediction without the sensors: Leveraging structure for unsupervised learning from monocular videos,'' in \emph{Proceedings of the AAAI conference on artificial intelligence}, 2019, pp. 8001--8008.

\bibitem{patil2020don}
V.~Patil, W.~Van~Gansbeke, D.~Dai, and L.~Van~Gool, ``Don’t forget the past: Recurrent depth estimation from monocular video,'' \emph{IEEE Robotics and Automation Letters}, vol.~5, no.~4, pp. 6813--6820, 2020.

\bibitem{wimbauer2021monorec}
F.~Wimbauer, N.~Yang, L.~Von~Stumberg, N.~Zeller, and D.~Cremers, ``Monorec: Semi-supervised dense reconstruction in dynamic environments from a single moving camera,'' in \emph{Proceedings of the IEEE/CVF Conference on Computer Vision and Pattern Recognition}, 2021, pp. 6112--6122.

\bibitem{gu2020cascade}
X.~Gu, Z.~Fan, S.~Zhu, Z.~Dai, F.~Tan, and P.~Tan, ``Cascade cost volume for high-resolution multi-view stereo and stereo matching,'' in \emph{Proceedings of the IEEE/CVF conference on computer vision and pattern recognition}, 2020, pp. 2495--2504.

\bibitem{klingner2020self}
M.~Klingner, J.-A. Term{\"o}hlen, J.~Mikolajczyk, and T.~Fingscheidt, ``Self-supervised monocular depth estimation: Solving the dynamic object problem by semantic guidance,'' in \emph{Computer Vision--ECCV 2020: 16th European Conference, Glasgow, UK, August 23--28, 2020, Proceedings, Part XX 16}.\hskip 1em plus 0.5em minus 0.4em\relax Springer, 2020, pp. 582--600.

\bibitem{feng2022disentangling}
Z.~Feng, L.~Yang, L.~Jing, H.~Wang, Y.~Tian, and B.~Li, ``Disentangling object motion and occlusion for unsupervised multi-frame monocular depth,'' in \emph{Computer Vision--ECCV 2022: 17th European Conference, Tel Aviv, Israel, October 23--27, 2022, Proceedings, Part XXXII}.\hskip 1em plus 0.5em minus 0.4em\relax Springer, 2022, pp. 228--244.

\bibitem{lee2021learning}
S.~Lee, S.~Im, S.~Lin, and I.~S. Kweon, ``Learning monocular depth in dynamic scenes via instance-aware projection consistency,'' in \emph{Proceedings of the AAAI Conference on Artificial Intelligence}, 2021, pp. 1863--1872.

\bibitem{wang2020deep}
J.~Wang, K.~Sun, T.~Cheng, B.~Jiang, C.~Deng, Y.~Zhao, D.~Liu, Y.~Mu, M.~Tan, X.~Wang, \emph{et~al.}, ``Deep high-resolution representation learning for visual recognition,'' \emph{IEEE transactions on pattern analysis and machine intelligence}, vol.~43, no.~10, pp. 3349--3364, 2020.

\bibitem{godard2017unsupervised}
C.~Godard, O.~Mac~Aodha, and G.~J. Brostow, ``Unsupervised monocular depth estimation with left-right consistency,'' in \emph{Proceedings of the IEEE conference on computer vision and pattern recognition}, 2017, pp. 270--279.

\bibitem{luo2019every}
C.~Luo, Z.~Yang, P.~Wang, Y.~Wang, W.~Xu, R.~Nevatia, and A.~Yuille, ``Every pixel counts++: Joint learning of geometry and motion with 3d holistic understanding,'' \emph{TPAMI}, vol.~42, no.~10, pp. 2624--2641, 2019.

\bibitem{shu2020feature}
C.~Shu, K.~Yu, Z.~Duan, and K.~Yang, ``Feature-metric loss for self-supervised learning of depth and egomotion,'' in \emph{ECCV}, 2020.

\bibitem{zhan2018unsupervised}
H.~Zhan, R.~Garg, C.~S. Weerasekera, K.~Li, H.~Agarwal, and I.~Reid, ``Unsupervised learning of monocular depth estimation and visual odometry with deep feature reconstruction,'' in \emph{CVPR}, 2018.

\bibitem{guizilini20203d}
V.~Guizilini, R.~Ambrus, S.~Pillai, A.~Raventos, and A.~Gaidon, ``3d packing for self-supervised monocular depth estimation,'' in \emph{Proceedings of the IEEE/CVF conference on computer vision and pattern recognition}, 2020, pp. 2485--2494.

\bibitem{godard2019digging}
C.~Godard, O.~Mac~Aodha, M.~Firman, and G.~J. Brostow, ``Digging into self-supervised monocular depth estimation,'' in \emph{Proceedings of the IEEE/CVF international conference on computer vision}, 2019, pp. 3828--3838.

\bibitem{ranjan2019competitive}
A.~Ranjan, V.~Jampani, L.~Balles, K.~Kim, D.~Sun, J.~Wulff, and M.~J. Black, ``Competitive collaboration: Joint unsupervised learning of depth, camera motion, optical flow and motion segmentation,'' in \emph{Proceedings of the IEEE/CVF conference on computer vision and pattern recognition}, 2019, pp. 12\,240--12\,249.

\bibitem{guizilini2020semantically}
V.~Guizilini, R.~Hou, J.~Li, R.~Ambrus, and A.~Gaidon, ``Semantically-guided representation learning for self-supervised monocular depth,'' \emph{Proceedings of the Eighth International Conference on Learning Representations}, 2020.

\bibitem{li2023learning}
R.~Li, D.~Xue, S.~Su, X.~He, Q.~Mao, Y.~Zhu, J.~Sun, and Y.~Zhang, ``Learning depth via leveraging semantics: Self-supervised monocular depth estimation with both implicit and explicit semantic guidance,'' \emph{Pattern Recognition}, vol. 137, p. 109297, 2023.

\bibitem{lee2023multi}
S.~Lee, W.~Im, and S.-E. Yoon, ``Multi-resolution distillation for self-supervised monocular depth estimation,'' \emph{Pattern Recognition Letters}, vol. 176, pp. 215--222, 2023.

\bibitem{bello2024self}
J.~L.~G. Bello, J.~Moon, and M.~Kim, ``Self-supervised monocular depth estimation with positional shift depth variance and adaptive disparity quantization,'' \emph{IEEE Transactions on Image Processing}, 2024.

\bibitem{zhang2024dcpi}
M.~Zhang, Y.~Feng, Q.~Chen, and R.~Fan, ``Dcpi-depth: Explicitly infusing dense correspondence prior to unsupervised monocular depth estimation,'' \emph{arXiv preprint arXiv:2405.16960}, 2024.

\bibitem{lee2021attentive}
S.~Lee, F.~Rameau, F.~Pan, and I.~S. Kweon, ``Attentive and contrastive learning for joint depth and motion field estimation,'' in \emph{Proceedings of the IEEE/CVF International Conference on Computer Vision}, 2021, pp. 4862--4871.

\bibitem{pilzer2018unsupervised}
A.~Pilzer, D.~Xu, M.~Puscas, E.~Ricci, and N.~Sebe, ``Unsupervised adversarial depth estimation using cycled generative networks,'' in \emph{2018 international conference on 3D vision (3DV)}.\hskip 1em plus 0.5em minus 0.4em\relax IEEE, 2018, pp. 587--595.

\bibitem{gordon2019depth}
A.~Gordon, H.~Li, R.~Jonschkowski, and A.~Angelova, ``Depth from videos in the wild: Unsupervised monocular depth learning from unknown cameras,'' in \emph{Proceedings of the IEEE/CVF International Conference on Computer Vision}, 2019, pp. 8977--8986.

\bibitem{li2021unsupervised}
H.~Li, A.~Gordon, H.~Zhao, V.~Casser, and A.~Angelova, ``Unsupervised monocular depth learning in dynamic scenes,'' in \emph{Conference on Robot Learning}.\hskip 1em plus 0.5em minus 0.4em\relax PMLR, 2021, pp. 1908--1917.

\bibitem{geiger2012we}
A.~Geiger, P.~Lenz, and R.~Urtasun, ``Are we ready for autonomous driving? the kitti vision benchmark suite,'' in \emph{CVPR}, 2012.

\bibitem{silberman2012indoor}
N.~Silberman, D.~Hoiem, P.~Kohli, and R.~Fergus, ``Indoor segmentation and support inference from rgbd images,'' in \emph{ECCV}, 2012.

\bibitem{he2016deep}
K.~He, X.~Zhang, S.~Ren, and J.~Sun, ``Deep residual learning for image recognition,'' in \emph{Proceedings of the IEEE conference on computer vision and pattern recognition}, 2016, pp. 770--778.

\bibitem{xu2022gmflow}
H.~Xu, J.~Zhang, J.~Cai, H.~Rezatofighi, and D.~Tao, ``Gmflow: Learning optical flow via global matching,'' in \emph{Proceedings of the IEEE/CVF conference on computer vision and pattern recognition}, 2022, pp. 8121--8130.

\bibitem{butler2012naturalistic}
D.~J. Butler, J.~Wulff, G.~B. Stanley, and M.~J. Black, ``A naturalistic open source movie for optical flow evaluation,'' in \emph{Computer Vision--ECCV 2012: 12th European Conference on Computer Vision, Florence, Italy, October 7-13, 2012, Proceedings, Part VI 12}.\hskip 1em plus 0.5em minus 0.4em\relax Springer, 2012, pp. 611--625.

\bibitem{jin2001real}
H.~Jin, P.~Favaro, and S.~Soatto, ``Real-time feature tracking and outlier rejection with changes in illumination,'' in \emph{ICCV}, 2001.

\bibitem{li2019pose}
Y.~Li, Y.~Ushiku, and T.~Harada, ``Pose graph optimization for unsupervised monocular visual odometry,'' in \emph{2019 International Conference on Robotics and Automation (ICRA)}.\hskip 1em plus 0.5em minus 0.4em\relax IEEE, 2019, pp. 5439--5445.

\end{thebibliography}
